\documentclass{bmvc2k}
\usepackage{amssymb} 
\usepackage{graphicx}
\usepackage{float}
\title{Future Does Matter: Boosting  3D Object Detection with Temporal Motion Estimation in Point Cloud Sequences}

\addauthor{Rui Yu}{y80220166@mail.ecust.edu.cn}{1}
\addauthor{Runkai Zhao}{rzha9419@uni.sydney.edu.au}{2}
\addauthor{Cong Nie}{2132586@tongji.edu.cn}{3}
\addauthor{Heng Wang}{hwan9147@uni.sydney.edu.au}{2}
\addauthor{HuaiCheng Yan}{hcyan@ecust.edu.cn}{1}
\addauthor{Meng Wang}{mengwagn@ecust.edu.cn}{1}

\addinstitution{East China University of Science and Technology, Shanghai, China
}
\addinstitution{University of Sydney, Sydney, Australia
}
\addinstitution{Tongji University,Shanghai, China
}

\runninghead{LiSTM}{Boosting  3D Object Detection with Temporal Motion Estimation}

\def\etal{\emph{et al}\bmvaOneDot}

\begin{document}

\maketitle

\begin{figure}[htbp]
\begin{tabular}{ccc}
\bmvaHangBox{\fcolorbox{white}{white}{\includegraphics[width=3.7cm]{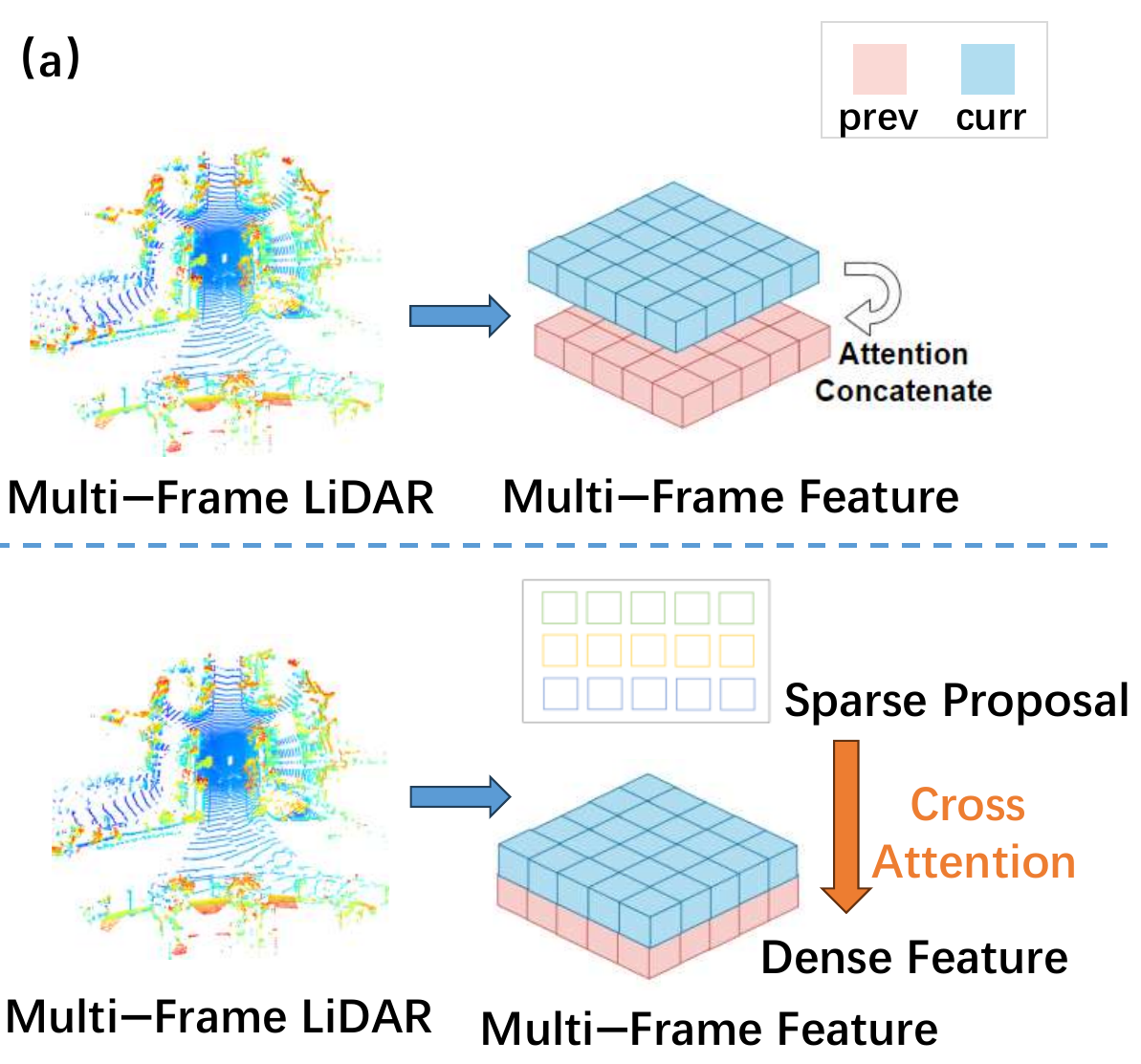}}}&
\bmvaHangBox{\fcolorbox{white}{white}{\includegraphics[width=3.5cm]{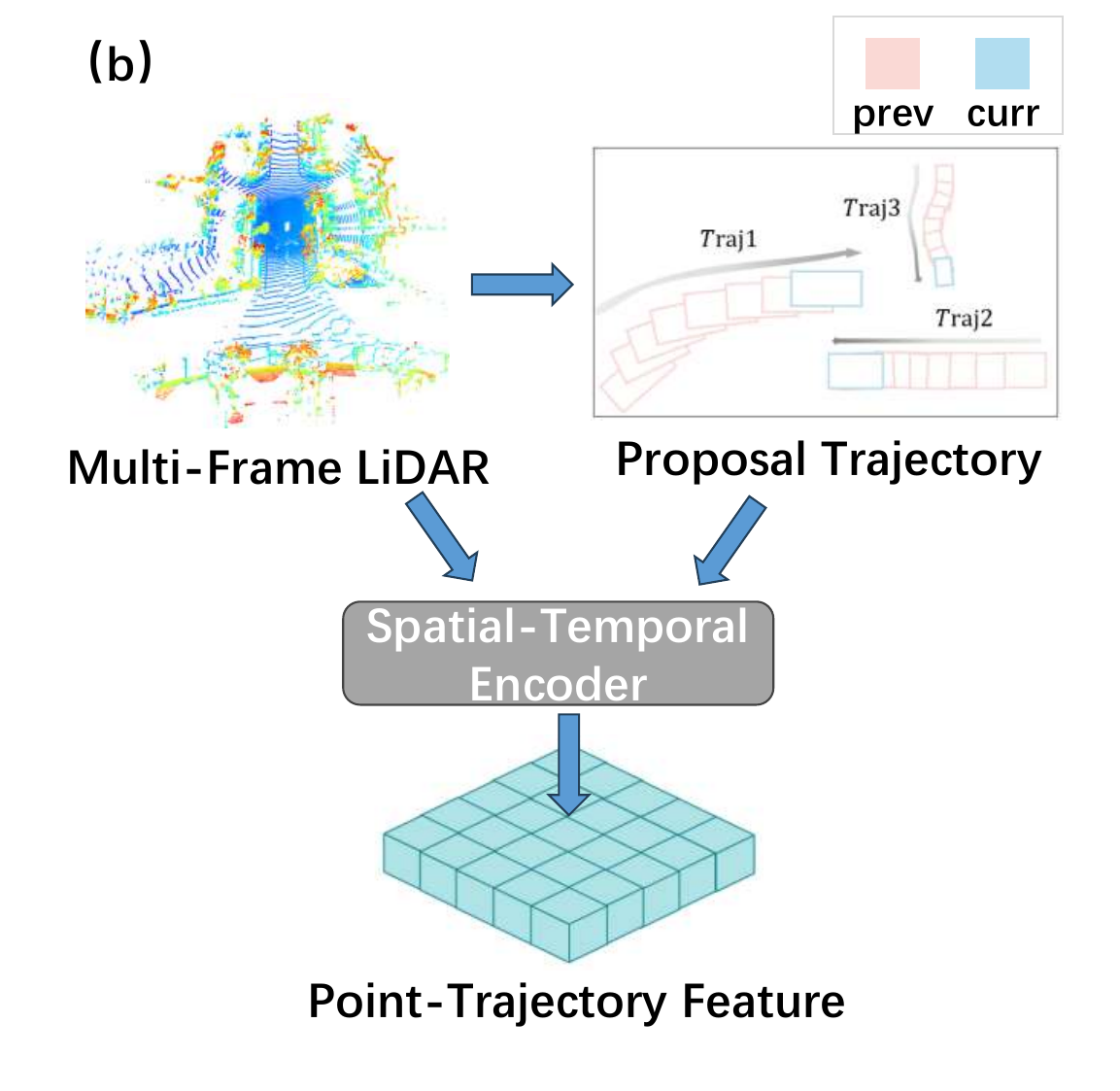}}}&
\bmvaHangBox{\fcolorbox{white}{white}{\includegraphics[width=4.2cm]{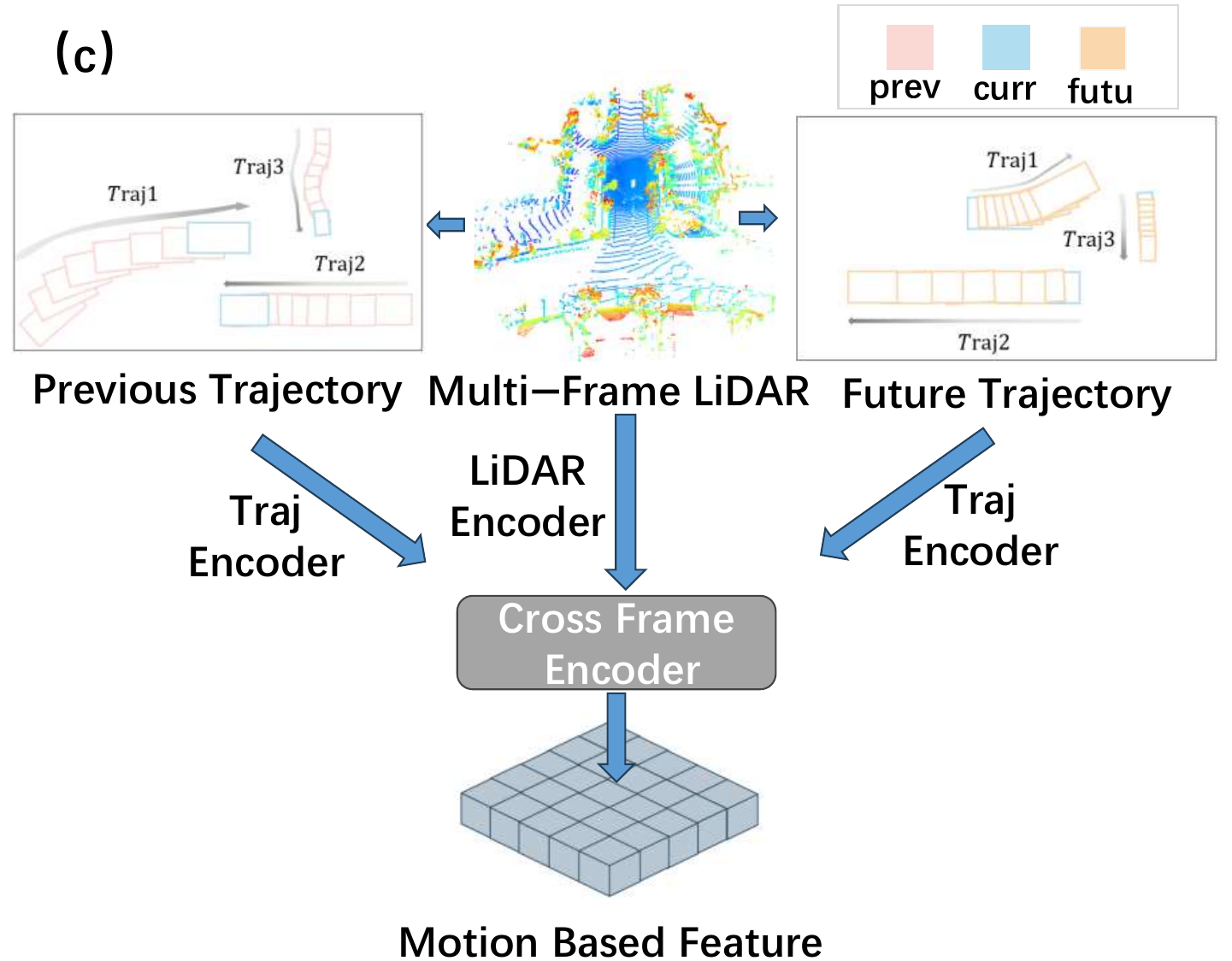}}}\\
\end{tabular}
\caption{Different from the global bird's eye view (BEV) Neighbor Feature Fusion Method (a) and Trajectory-based Method (b) which do not count for the role of the future states, we propose a novel LiDAR 3D object detection framework that utilizes motion forecasting to guide the temporal fusion learning across past and future frames as shown in (c).}
\label{fig:fig1}
\end{figure}

\begin{abstract}
Accurate and robust LiDAR 3D object detection is essential for comprehensive scene understanding in autonomous driving. Despite its importance, LiDAR detection performance is limited by inherent constraints of point cloud data, particularly under conditions of extended distances and occlusions. Recently, temporal aggregation has been proven to significantly enhance detection accuracy by fusing multi-frame viewpoint information and enriching the spatial representation of objects. In this work, we introduce a novel LiDAR 3D object detection framework, namely \textit{\textbf{LiSTM}}, to facilitate spatial-temporal feature learning with cross-frame motion forecasting information. We aim to improve the spatial-temporal interpretation capabilities of the LiDAR detector by incorporating a dynamic prior, generated from a non-learnable motion estimation model. Specifically, Motion-Guided Feature Aggregation (MGFA) is proposed to utilize the object trajectory from previous and future motion states to model spatial-temporal correlations into gaussian heatmap over a driving sequence. This motion-based heatmap then guides the temporal feature fusion, enriching the proposed object features. Moreover, we design a Dual Correlation Weighting Module (DCWM) that effectively facilitates the interaction between past and prospective frames through scene- and channel-wise feature abstraction. In the end, a cascade cross-attention-based decoder is employed to refine the 3D prediction. We have conducted experiments on the Waymo and nuScenes datasets to demonstrate that the proposed framework achieves superior 3D detection performance with effective spatial-temporal feature learning. \href{https://github.com/YuRui-Learning/LiSTM}{https://github.com/YuRui-Learning/LiSTM}
\end{abstract}

\section{Introduction}
\label{sec:intro}

3D LiDAR object detector \cite{pointpillars,voxelnet,centerpoint} plays an important role in autonomous driving, it identifies object information within a 3D road scene represented by an unstructured point cloud. Although discrete LiDAR points reflect accurate spatial positioning of surrounding driving scenes, they are insufficient to comprehensively describe traffic objects due to data sparsity, particularly at far distances. Moreover, the LiDAR sensor captures partial view information of a scene from a single-frame perspective, leading to incomplete information collection of the visible objects. These inherent limitations of LiADR result in inconsistent point distribution for the same object across a driving sequence. Hence, a dynamic object may be represented with varying densities of point clouds in different frames, which introduces ambiguity in accurately determining the true shape for a 3D detector.

To eliminate the inconsistency, the increasing works \cite{centerformer,timepillars,convgru} attempt to detect 3D objects by utilizing multiple frames of point clouds. 
The LiDAR sensor records driving scenarios as the vehicle moves, delineating objects across multiple perspectives in sequence. This adds valuable modal information, enriching object representation. 
A straightforward method to implement this idea is to fuse the neighboring frame features, using the insight of historical frames to enhance the semantic representation of the current scene. Referring to the application of transformer in computer vision, the cross-attention mechanism bridges the previous and current point features either densely or sparsely, as depicted in Figure \ref{fig:fig1}(a).

Direct integration of features for historical frames enhances the detection performance, but this method struggles to handle fast-moving objects. To solve this issue, trajectory-based methods \cite{mppnet,he2023msf} are designed to aggregate extensive temporal contexts of the object flows and utilize multi-frame proposals to comprehend the spatial information among the driving scenes. As shown in Figure \ref{fig:fig1}(b), this method enhances the representation of the object by incorporating multi-view complementary information from the corresponding trajectory. However, this input-level manipulation is resource-intensive, limiting detection efficiency.

To boost temporal object detection, we propose a novel \textbf{Li}DAR 3D object detection with enhancing \textbf{S}patial-\textbf{T}emporal feature fusion through  \textbf{M}otion estimation, namely \textbf{\textit{LiSTM}}. We bolster spatial-temporal feature fusion by integrating a Kalman filter module \cite{eagermot} as prior kinetic information and focus on effectively integrating both ego and object motion states. 
Unlike previous approaches \cite{mppnet,he2023msf} that directly encode proposal trajectories with point clouds, we uncover an implicit feature representation for both trajectories and point clouds within the BEV space using the motion-based heatmap generator. This enables direct feature-level fusion, eliminating the need for reliance on the PointNet \cite{pointnet} backbone. To have a stronger dynamic prior for each frame, we design the \textbf{M}otion-\textbf{G}uided \textbf{F}eature \textbf{A}ggregation (MGFA) mechanism to combine the heatmap generated by trajectory prediction for guiding the reconstruction of LiDAR features. Ultimately, with the integration of the \textbf{D}ual \textbf{C}orrelation \textbf{W}eighting \textbf{M}odule (DCWM) and Motion Transformer, we enhance feature characterization across frames, thereby enriching the semantic and geometric representations.

The main contributions of this paper can be summarized as follows:
\begin{itemize}
\item We propose a novel LiDAR object detector considering future motion estimation of objects and point clouds to enhance the effectiveness of the spatial-temporal fusion.
\item We design a Motion-Guided Feature Aggregation (MGFA) mechanism to enhance object geometric representations of motions, and the Dual Correlation Weighting Module (DCWM) to characterize the spatial relationship of features across sequences.
\item We conduct experiments on the nuScenes and Waymo datasets to validate our proposed framework, which outperforms  CenterPoint by 8\% on the Waymo dataset.
\end{itemize}


\section{Related Work}
\label{sec:intro}
\textbf{BEV 3D Object Detection.}
The bird's-eye view (BEV) is a widely used feature representation in the field of autonomous driving which is derived from LiDAR's ability to perceive objects from a circular viewpoint. Thanks to the PointNet series \cite{pointnet}, point-based methods \cite{pointrcnn} have become extensively employed to extract geometric features directly from point clouds. Voxel-based methods \cite{voxelnet,voxelnext,3dlane} and Pillar-based methods \cite{pointpillars,pillarnet} are mainly applied in environmental perception by converting point cloud to BEV feature. 
Meanwhile, Camera-based detectors \cite{lss,bevdepth} learn pixel-wise categorical depth distributions to lift  2D images of different views into BEV space. Additionally, Li \etal~ \cite{bevformer} proposes a spatiotemporal transformer and focus on feature fusion in the spatial-temporal 4D working space.   

\noindent\textbf{Keypoint Detection.}
Anchor-based methods \cite{yolov3,ssd} often result in redundant bounding boxes, requiring the use of Non-Maximum Suppression. Law and Deng \cite{cornnet} produce two corner pairs to detect, while Zhou \etal~ \cite{objectspoint} uses keypoint estimation with a normal distribution to locate center points, which use the central region to regress other properties. Therefore, CenterPoint \cite{centerpoint} follows the struct of CenterNet \cite{objectspoint} and employs an object detector in BEV space. Zhou \etal~ \cite{centerformer} utilizes the initial query embedding to facilitate learning of the transformer and uses cross attention to efficiently aggregate neighboring features. 

\noindent\textbf{Temporal Fusion Methodology.}
Temporal Fusion plays a critical role in autonomous driving, allowing models to gain a deeper understanding of contextual geometric information. Zhou \etal~ \cite{centerformer} performs multi-frame features fusion by utilizing spatial-aware attention, while RNN-based models \cite{timepillars,convgru} employ LSTM and GRU to fuse previous state features with the current feature. BEVFormer \cite{bevformer} designs a temporal deformable attention to fuse previous features for enhanced performance. Meanwhile, Wang \etal~ \cite{streampetr} develops an object-centric temporal mechanism and a motion-aware layer normalization to model the movement of the objects. 3D-MAN \cite{3dman} utilizes a multi-frame alignment and aggregation module to learn temporal attention for detection from multiple frames. motion-based models \cite{mppnet,he2023msf,ptt} design point-trajectory transformer with long short-term memory for efficient temporal 3D object detection. Li \etal~ \cite{modar} uses motion forecasting outputs as a type of virtual lightweight sensor modality. Hence, we propose a more powerful and efficient spatial-temporal fusion model under BEV using CenterPoint \cite{centerpoint} as the baseline.

\begin{figure}[h]
    \centering
    \includegraphics[width=0.75\textwidth]{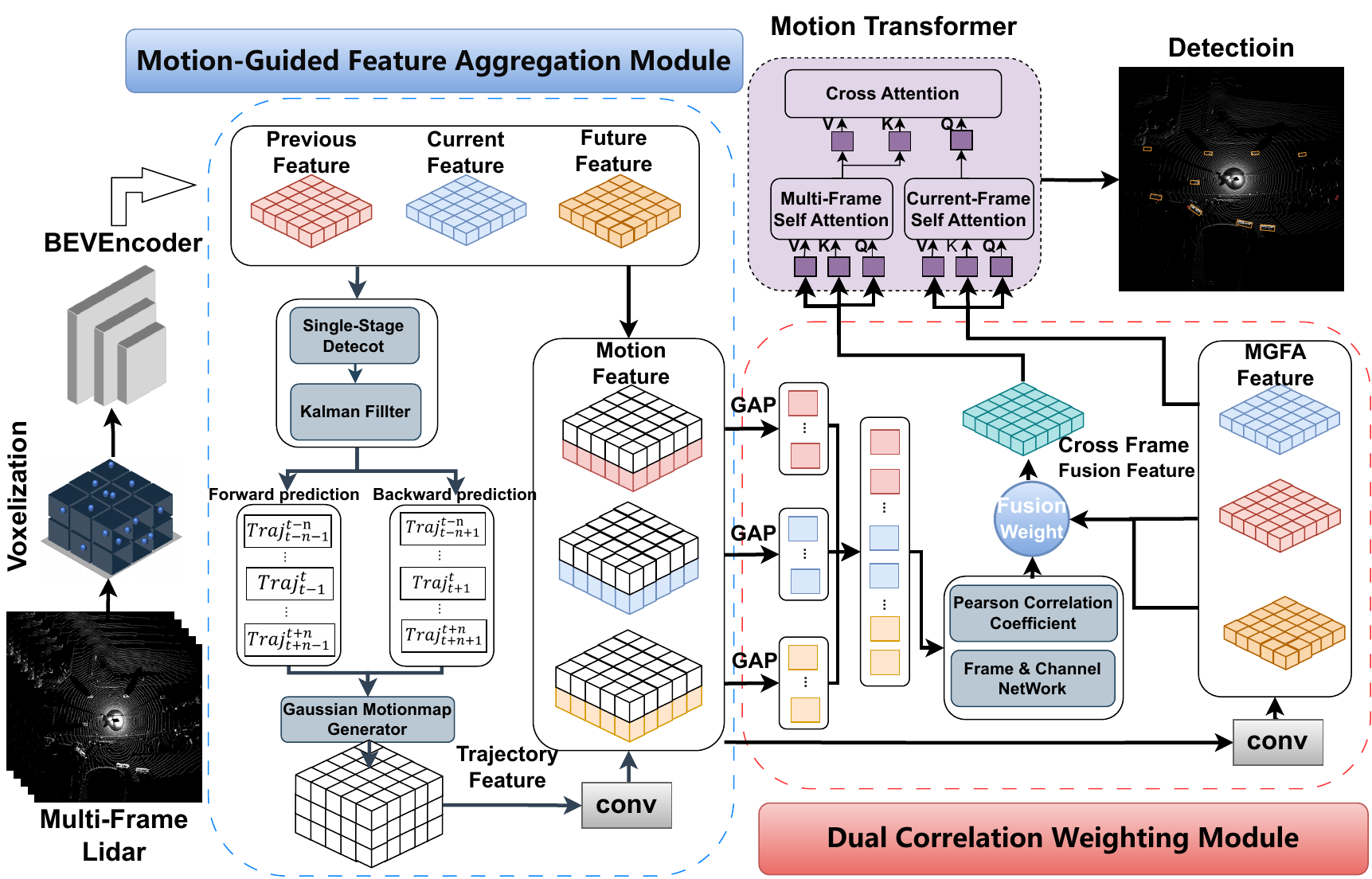}
    \caption{Overview of our proposed framework \textit{\textbf{LiSTM}}. It processes multi-frame point clouds by performing voxelization before feeding them into the LiDAR BEV encoder. The first module employs a single-stage detector combined with tracking prediction to produce trajectories and then enhances the spatial representation with a Motion-Guided Feature Aggregation Module. The second module is used for cross-frame feature extraction by the proposed Dual Correlation Weighting Module and Motion Transformer. }
    \label{fig:fig2}
\end{figure}

\section{Approach}
As depicted in Figure \ref{fig:fig2}, to incorporate the motion prior, we focus in Section \ref{3.1} on the generation of the motion feature and the Motion-Guided Feature Aggregation (MGFA) mechanism. Then, the Dual Correlation Weighting Module (DCWM) and the Motion Transformer will be presented in Sections \ref{3.3} and \ref{3.4} to describe the cross-frame fusion strategy.

\subsection{Motion-Guided Feature Aggregation }
\label{3.1}
Unlike the early fusion methods \cite{mppnet,he2023msf}, we utilize motion-based heatmap representing temporal streams to normalize features of objects for deep fusion. 
To predict object positions in future scenes, we use a kinematic model of ego-motion to derive the transformation matrix from time $t$ to $t + n$, based on prior motion data and ego-pose observations.
The transformation matrix is then used to transfer the point cloud in the current scene to a future coordinate, but it only applies to static objects and obtains a coarse-grained prediction.
However, whether the points are predictions or observations are processed through voxelization and encoder to produce features $F_{multi} = \{F_{t-n},...,F_{t+n}\}$. Then as implemented in CenterPoint \cite{centerpoint}, we can get multi-frame proposals, which are temporal independence and geometric correlation.

\noindent\textbf{Motion Model.}
After acquiring multiple consecutive frames of object proposals, we can use a Kalman filtering \cite{eagermot} to estimate the motion state of each object across the frames. We define a ten-dimension state space $(x,y,z,\theta,l,w,h,\dot{x},\dot{y},\dot{z})$, where $B=(x,y,z)$ is the center of a 3d bounding box, $P_{dim}=(l,w,h)$ is the object size, $\theta$ is the orientation under BEV and $V=(\dot{x},\dot{y},\dot{z})$ are the respective velocities in the 3D space learned by a Kalman filter for constant velocity motion with a linear observation model.

\noindent\textbf{Trajectory Prediction.}
With the Kalman filter modeling multiple targets over a driving sequence, we can obtain information about the velocity prediction $V^{t}$ of each proposal at every moment. 
For the forward trajectory prediction, we utilize the bounding box observation $B_{t-1}$ at $t-1$, along with the velocity prediction $V_{t-1}$, to update the $B'{t}$ for frame $t$. Similarly, for the reverse trajectory prediction, we employ the bounding box observation $B_{t+1}$ at $t+1$ and the updated velocity prediction $V_{t+1}$ to reverse-predict the predicted the $B''_{t}$:
\begin{align}
B'_{t} = B_{t-1} + V^{t-1} \cdot \Delta t, \\
B''_{t} = B_{t+1} - V^{t+1} \cdot \Delta t.
\end{align}
\noindent\textbf{Motion-based Heatmap Generator.}
After acquiring the forward and backward trajectory predictions $B'_{t}$ and $B''_{t}$, we transfer these trajectories into motion feature $F_{motion}$ using gaussian distribution. As is known, gaussian distribution is determined as:
\begin{equation}
\mu_{k}^{x}=cx_{k}, \quad \mu_{k}^{y}=cy_{k},
\end{equation}
\noindent where $\mu_{k}$ represents the location of the proposal under BEV, and $\sigma_{k}$ is the hyperparameter of the category associated with the category of the $k^{th}$ object.

For the normal representations $N_{t-1}^{t}(\mu_{k},\sigma_{k}^{2})$ and $N_{t+1}^{t}(\mu_{k},\sigma_{k}^{2})$ of each frame proposal generated by bidirectional trajectory prediction, We respectively use the $\sigma_{k}$ to control the probability of the distribution and the $\mu_{k}$ to represent the center of the distribution. 
Given the proposals from neighboring frames, we can consolidate all distributions into the BEV representation \( F_{\text{motion}} \), which enhances the understanding of agent objects by providing additional motion modality insights.
This can be very effective in solving fast-moving objects and supplying a prior for occlusion situations.

\noindent\textbf{Motion Guided Feature Aggregation Module.}
The designed MGFA module utilizes the information from previous and future motion states to interact with dense BEV features to model spatial-temporal correlations. By incorporating $F_{motion}$,  we can enrich positional semantic information and integrate motion characterization into the model's understanding. 
As mentioned, the motion feature includes bidirectional projections for the target frame. Therefore, the specific motion features are denoted as follows, with \( p2c \) and \( f2c \) representing past and future predictions of the current frame, respectively:
\begin{align}
F_{motion}^{p2c}=\{{N_{t-n-1}^{t-n}},...N_{t-1}^{t},...,N_{t+n-1}^{t+n}\}, \\
F_{motion}^{f2c}=\{{N_{t-n+1}^{t-n}},...N_{t+1}^{t},...,N_{t+n+1}^{t+n}\}.
\end{align}

\begin{figure}[htbp]
    \begin{minipage}{0.52\textwidth}
        \bmvaHangBox{\fcolorbox{white}{white}{\includegraphics[width=5.6cm]{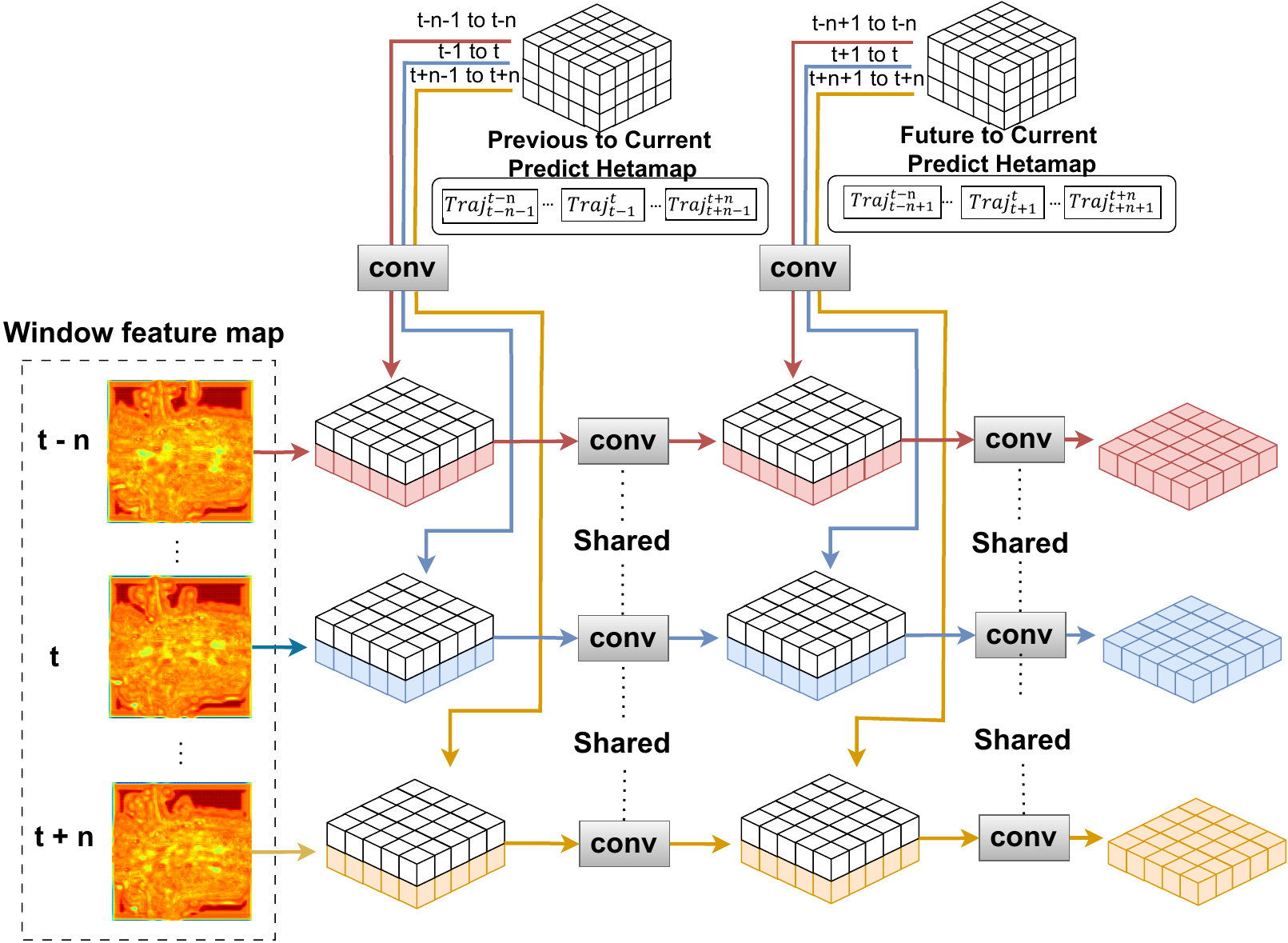}}}
        \caption{Motion Guided Feature Aggregation.}
        \label{fig:fig3}
    \end{minipage}\hfill 
    \begin{minipage}{0.53\textwidth}
        \bmvaHangBox{\fcolorbox{white}{white}{\includegraphics[width=6.4cm]{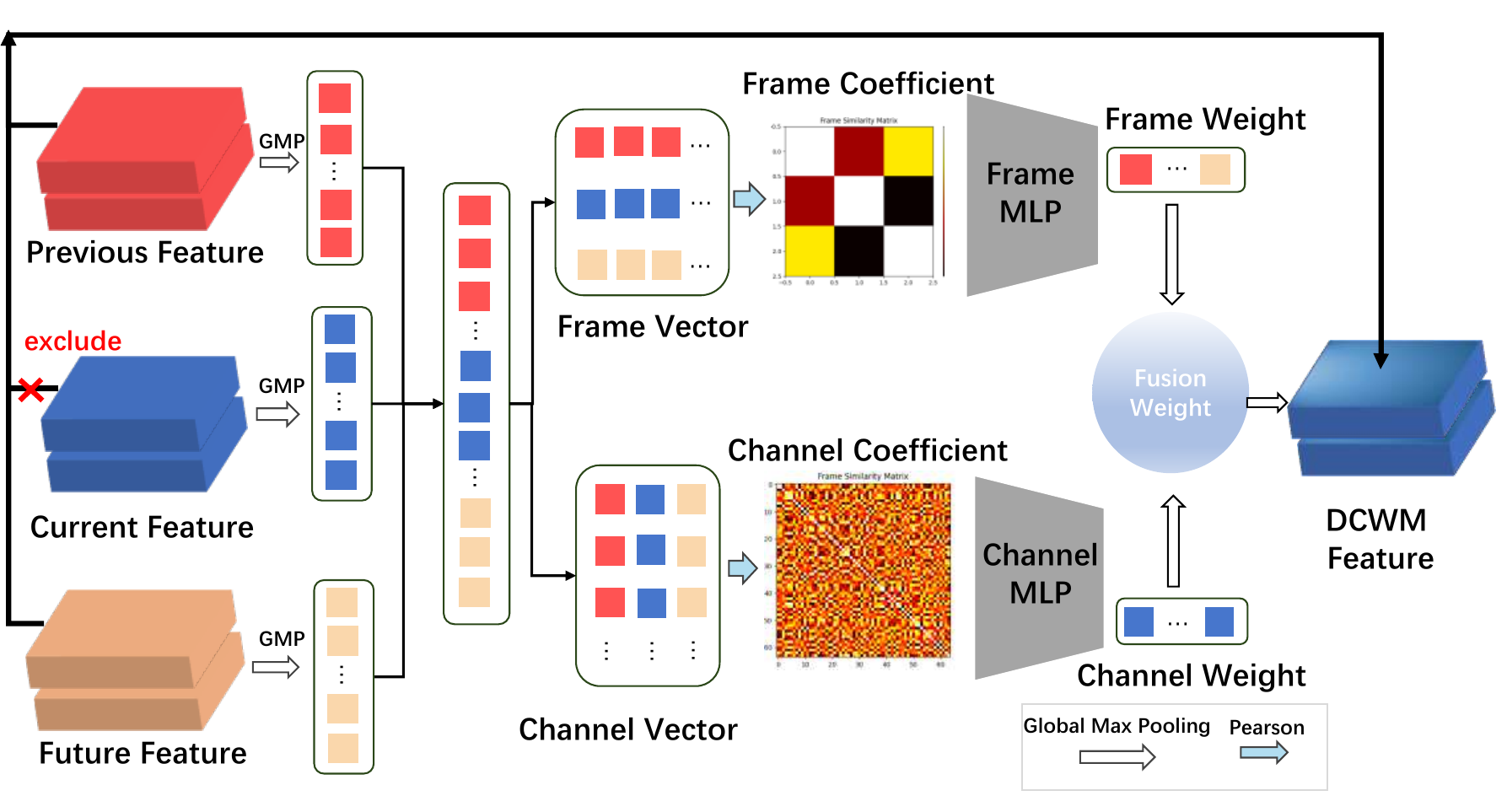}}}
        \caption{Dual Correlation Weighting Module.}
        \label{fig:fig4}
    \end{minipage}
\end{figure}

Based on the given feature $F_{motion}$, it is first expanded along the channel dimension and then processed by a shared $Conv$ to encode the geometric information of the target center. Specifically, $Conv$ denotes channel expansion followed by dimensionality reduction within the channel dimension:

\begin{equation}
F^{p2c /f2c}_{center}=Conv(repeat(F_{motion}^{p2c/f2c})).
\end{equation}

After obtaining the center distribution feature $F^{p2c /f2c}_{center}$, we follow the method illustrated in Figure \ref{fig:fig3} to perform the
feature fusion using a shared convolutional network. 
In the aggregation of forward prediction, we merge the forward distribution feature $F^{p2c}_{center}$ of the target frame from the previous frame with the BEV feature by using a convolutional network. It mainly convolves the channel dimension to realize the fusion of heterogeneous features:
\begin{equation} 
F_{MGFA}=Conv(F_{multi},F^{p2c }_{center})=\{Conv(F_{t-n},N_{t-n-1}^{t-n}),...,Conv(F_{t+n},N_{t+n-1}^{t-n})\}.
\end{equation}

Similarly, in reverse trajectory prediction, the center distribution feature $F^{f2c }_{center}$ is sequentially concatenated and convolved with the BEV feature to enhance the dynamic property:
\begin{equation}
F'_{MGFA}=Conv(F_{MGFA},M^{f2c }_{center})=\{Conv(F'_{t-n},N_{t-n+1}^{t-n}),...,Conv(F'_{t+n},N_{t+n+1}^{t-n})\}.
\end{equation}

\subsection{Dual Correlation Weighting Module  }
\label{3.3}

Unlike the feature concatenation in Figure \ref{fig:fig1}(a), we propose learning a  multi-frame fusion weight matrix to capture cross-frame correlations in both channel and temporal dimensions.
As shown in Figure \ref{fig:fig4}, global max pooling ($GMP$) is first applied along the spatial dimensions to obtain a feature vector $v_{t}$. Subsequently, vectors from multiple frames are concatenated to form a representation for the scene sequence data, denoted as $V=\{v_{t-n},...,v_{t+n}\}$:

\begin{equation}
V=Concat(v_{t-n},..,v_{t+n}) = Concat(GMP\{F^{'t-n}_{MGFA}\},..,GMP\{F_{MGFA}^{'t+n}\}),
\end{equation}
\begin{equation}
M_{d/t}=\frac{conv(V^{d/t}_{i},V^{d/t}_{j})}{\sigma_{V^{d/t}_{i}}*\sigma_{V^{d/t}_{j}}}.
\end{equation}

We then compute the correlation between matrices across each vector(e.g., i and j), where $V^{d/t}$ denotes the process of transforming the sequence along the channel and temporal. 
After obtaining the correlation matrices $M_{d}$ and $M_{t}$, which represent interlinks within the feature structure and across frames in the temporal domain, respectively, the weight matrix is flattened and passed through a two-layer linear network with ReLU activation:
\begin{equation}
W_{d/t}=Linear(ReLu(Linear(M_{d/t}))).
\end{equation}

Eventually, we obtain weight vectors $W^{d/t}$ for channels and temporal dimensions, respectively, and generate the weight matrix \( M_{\text{weight}} \) through their outer product $\otimes$.
Then, this weight is multiplied and channel-wise convolution with the MGFA feature $F''_{MGFA}$ (excluding the current frame) to generate the Dual Correlation Weighting feature $F_{DCWM}$ as follows:
\begin{equation}
F_{DCWM}=Conv(F''_{MGFA}\cdot M_{weight})=Conv(F''_{MGFA} \cdot (W_{d} \otimes W_{t})).
\end{equation}

\subsection{Motion Transformer}
\label{3.4}
With the assistance of the designed modules MGFA and DCWM, the features are enhanced to include details about both ego-motion and object-motion.
The attention mechanism \cite{attention} is then employed using a transformer decoder to focus on feature learning within the spatial-temporal 4D space.
First, the features are processed through self-attention as follows:
\begin{equation}
Q_{C/M}=MultiHeadAttn(Q(F_{C/M}+PE),K(F_{C/M}+PE),V(F_{C/M})),
\end{equation}

\noindent where $F_{C/M}$ represents the current frame feature and DCWM feature, $Q_{C/M}$ denotes the query and PE is the position embedding.
After self-attention, we make a cross-attention mechanism with $Q_{C}$ and $Q_{M}$, which guides the training to focus on aggregating more spatial information containing meaningful object details. 
Then, the cross-attention is shown below:
\begin{equation}
Q_C'=MultiHeadAttn(Q(Q_{C}+PE),K(Q_{M}+PE),V(Q_{M})).
\end{equation}
After feature generation and fusion, we get the final target characterization $Q_C'$. Then we follow the steps of CenterPoint \cite{centerpoint} to learn the representation of the different geometric elements in the 3D scene. 

\begin{table}
\centering
\fontsize{6.6}{7.5}\selectfont
\begin{tabular}{@{\extracolsep{\fill}}l|c|cc|cc|cc}
\hline
\textbf{Model} & \textbf{Frames} & \multicolumn{2}{c|}{\textbf{Vehicle (AP/APH)$\uparrow$}} & \multicolumn{2}{c|}{\textbf{Pedestrian (AP/APH)$\uparrow$}} & \multicolumn{2}{c}{\textbf{Cyclist (AP/APH)$\uparrow$}} \\
 & & \textbf{L1} & \textbf{L2} & \textbf{L1} & \textbf{L2} & \textbf{L1} & \textbf{L2} \\
\hline
PointPillar \cite{pointpillars} & 1 & 66.94 / 66.36 & 58.96 / 58.43 & 63.35 / 45.22 & 55.21 / 39.32 & 55.06 / 52.55 & 52.97 / 50.55 \\
VoxelNet \cite{voxelnet} & 1 & 68.73 / 67.31 & 60.11 / 59.97 & 69.65 / 57.38 & 60.19 / 53.67 & 62.31 / 59.85 & 60.34 / 55.89 \\
PillarNet \cite{pillarnet} & 1 & 66.29 / 65.63 & 59.03 / 58.43 & 70.35 / 64.24 & 64.24 / 55.75 & 65.43 / 63.93 & 63.53 / 62.08 \\
Second \cite{second}& 1 & 68.95 / 68.33 & 61.81 / 61.24 & 65.59 / 54.80 & 57.85 / 48.16 & 61.14 / 59.50 & 56.84 / 55.26 \\
CenterPoint \cite{centerpoint} & 1 & 67.87 / 67.27 & 59.96 / 59.43 & 69.31 / 62.55 & 61.17 / 55.06 & 64.28 / 63.05 & 61.86 / 60.68 \\
\hline
PartA2 \cite{PARTA2} & 1 & 65.52 / 64.85 & 57.32 / 56.63 & 54.83 / 37.72 & 46.85 / 32.19 & 54.29 / 48.75 & 52.21 / 46.89 \\
PVRCNN \cite{pvrcnn} & 1 & 71.11 / 70.32 & 62.60 / 61.88 & 63.63 / 32.77 & 54.88 / 28.26 & 59.49 / 34.14 & 57.22 / 32.83 \\
VoxelRCNN \cite{voxelrcnn} & 1 & 71.51 / 70.98 & 63.75 / 63.26 & 65.95 / 65.99 & 65.47 / 60.86 & 70.11 / 68.71 & 67.98 / 66.63 \\
\hline
CenterPoint \cite{centerpoint} & 4 & 71.27 / 70.73 & 63.59 / 63.09 & 73.91 / 70.45 & 66.28 / 60.10 & 63.78 / 62.98 & 61.59 / 60.82 \\
CenterPoint \cite{centerpoint} & 16 & 72.53 / 71.31 & 64.18 / 64.21 & 74.05 / 71.17 & 66.17 / 61.03 & 64.05/ 64.54 & 62.31 / 61.77 \\
MPPNet \cite{mppnet} & 4 & 74.24 / 73.55 & 66.29 / 65.38 & \underline{76.94} / \underline{72.29} & \underline{68.63} / \underline{66.16} & 67.34/ 66.67 & 65.12 / 64.48 \\
MSF \cite{he2023msf} & 4 & \underline{74.37} / \underline{73.97} &  \underline{66.35} / \underline{65.85} & \bfseries 78.16 / 74.91 & \bfseries 70.27 / 67.21 & \underline{67.89}/ \underline{67.14} & \underline{65.58} / \underline{64.89} \\
LiSTM & 3 &\bfseries 74.83 / 74.32 & \bfseries 66.85 / 66.17 &  75.89 / 69.72 & 66.83 / 63.43 & \bfseries 70.84 / 69.75 & \bfseries 68.23 / 69.12 \\
\hline
\end{tabular}
\caption{Quantative comparisons on 20\% Sequence Waymo validation set.}
\label{tab:1}
\end{table}

\noindent\begin{table}
\centering
\fontsize{7.5}{8.5}\selectfont
\begin{tabular}{@{}lccccccc@{}}
\hline
\textbf{Model} & \textbf{NDS$\uparrow$} & \textbf{mAP$\uparrow$} & \textbf{mATE$\downarrow$} & \textbf{mASE$\downarrow$} & \textbf{mAOE$\downarrow$} & \textbf{mAVE$\downarrow$} & \textbf{mAAE$\downarrow$} \\
\hline
PointPillar \cite{pointpillars} & 58.62 & 45.27 & 0.3353 & 0.259 & 03286 & 0.2784 & 0.2002 \\
Second \cite{second} & 62.31 & 50.8 & 0.3140 & 0.2554 & \bfseries{0.2785} & 0.2587 & 0.2019 \\
CenterPoint \cite{centerpoint} & 66.29 & 58.77 & \underline{0.2919} & 0.2566 & 0.3692 & \bfseries{0.2081} & \bfseries 0.1837 \\
VoxelNext \cite{voxelnext} & \underline{67.09} & \underline{60.55} & 0.3023 & \underline{0.2526} & 0.3701 & \underline{0.2087} & 0.1851 \\
LiSTM & \bfseries 68.32 & \bfseries 63.77 & \bfseries 0.2895 & \bfseries 0.2479 & \underline{0.3182} & 0.2472 & \underline{0.1850} \\
\hline
\end{tabular}
\caption{Quantative comparisons on nuScenes validation set.}
\label{tab:2}
\end{table}

\section{Experiments}
\label{sec:intro}
\noindent\textbf{Dataset and Metrics.}
The Waymo Open dataset \cite{waymodataset} is a highly regarded benchmark for automatic driving. It consists of 1150 point cloud sequences, with over 200,000 frames in total. Evaluation of results using mean Average Precision (mAP) and its weighted variant by heading accuracy (mAPH). Results are reported for LEVEL 1 (L1, easy only) and LEVEL 2 (L2, easy and hard) difficulty levels, considering vehicles, pedestrians, and cyclists.

The nuScenes dataset \cite{nuscenes} provides diverse annotations for autonomous driving and features challenging evaluation metrics. These include mean Average Precision (mAP) at four center distance thresholds and five true-positive metrics: ATE, ASE, AOE, AVE, and AAE, which measure translation, scale, orientation, velocity, and attribute errors, respectively. Additionally, the nuScenes detection score (NDS) combines mAP with these metrics.

\noindent\textbf{Experimental Settings.}
In our experimental setup, we follow the default settings of Openpcdet \cite{openpcdet} and conduct the experiments using two 24GB Nvidia RTX 3090 GPUs. The validation process utilized the nuScenes and Waymo datasets. We employed the AdamW optimizer with a base learning rate of \(3 \times 10^{-3}\) and applied layer-wise learning rate decay.

\noindent\textbf{Comparison Experiment.}
We validate the effectiveness of the designed LiSTM on Waymo's validation set (Table \ref{tab:1}), using 20\% of the sequences for training.
Full results are available in Table \ref{tab:8} of the Appendix.
LiSTM achieves an impressive improvement of over 8\% compared to single-stage models like CenterPoint \cite{centerpoint}, while also outperforming two-stage models such as PVRCNN \cite{pvrcnn} and VoxelRCNN \cite{voxelrcnn}.
Meanwhile, LiSTM, a multi-frame single-stage model, eliminates the need for region-of-interest extraction, resulting in reduced resource consumption, as illustrated in Table \ref{tab:7}.
In comparison to multi-frame CenterPoint \cite{centerpoint}, LiSTM achieves remarkable improvements while utilizing fewer frames. When compared to two-stage models MPPNet \cite{mppnet} and MSF \cite{he2023msf}, LiSTM demonstrates clear advancements in vehicle and cyclist detection which is attributed to motion-based feature integration. More details and discussions can be found in the Appendix.

In Figure \ref{fig:fig5}, we compare the baseline with our module. LiSTM demonstrates superior capability, particularly highlighted by the pink arrows, in detecting cases that CenterPoint fails to identify due to distance and occlusion challenges. 
Additionally, LiSTM offers an increased number of positive samples with no annotations, as indicated by the yellow arrows.

On the nuScenes dataset, LiSTM outperforms the benchmarks PointPillar \cite{pointpillars} and VoxelNet \cite{voxelnet}, improving NDS and mAP by 2-3\% compared to CenterPoint \cite{centerpoint}. Meanwhile, LiSTM is a boost in ATE and ASE  as noted in Table \ref{tab:2}.

\begin{figure}[htbp]
\begin{tabular}{ccc}
\bmvaHangBox{\fcolorbox{white}{white}{\includegraphics[width=3.7cm]{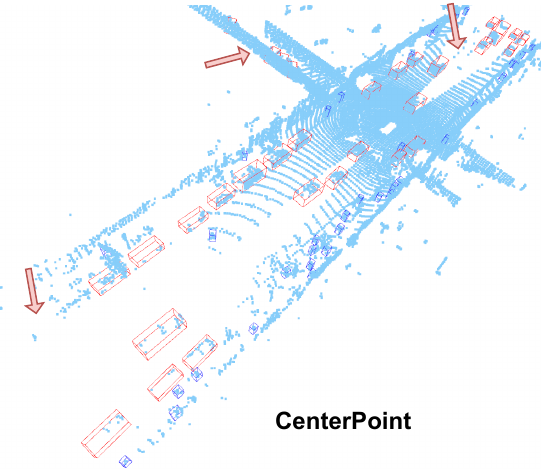}}}&
\bmvaHangBox{\fcolorbox{white}{white}{\includegraphics[width=3.7cm]{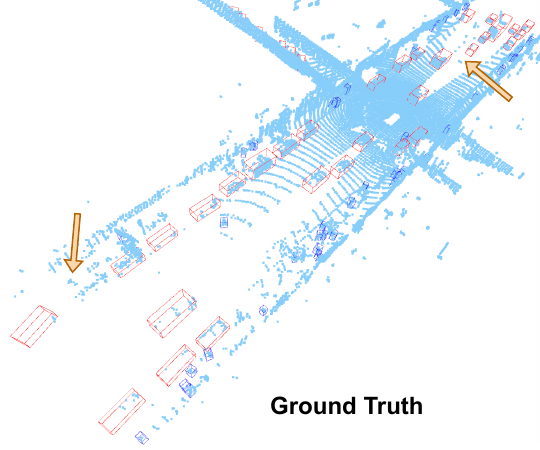}}}&
\bmvaHangBox{\fcolorbox{white}{white}{\includegraphics[width=3.7cm]{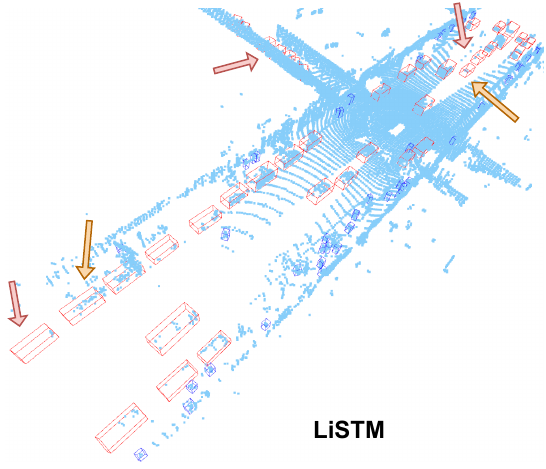}}}\\
\end{tabular}
\caption{Qualitative visualization of our LiSTM on Waymo validation set. We show the 3D boxes predictions in the LiDAR bird's-eye-view}
\label{fig:fig5}
\end{figure}

\begin{table}
\centering
\fontsize{7.5}{8.5}\selectfont
\begin{tabular}{cccc|ccc}
\hline
\textbf{CenterPoint} & \textbf{MotionTransformer} & \textbf{MGFA} & \textbf{DCWM} & \textbf{Veh. L2 APH} & \textbf{Ped. L2 APH} & \textbf{Cyl. L2 APH} \\
\hline
$\checkmark$ & $\times$ & $\times$ & $\times$ & 59.51 & 55.22 & 60.54 \\
$\checkmark$ & $\checkmark$ & $\times$ & $\times$ & 62.49 & 56.04 & 63.17 \\
$\checkmark$ & $\checkmark$ & $\checkmark$ & $\times$ & 64.67 & 57.56 & 67.86 \\
$\checkmark$ & $\checkmark$ & $\checkmark$ & $\checkmark$ & \bfseries 65.88 & \bfseries 61.10 &  \bfseries 68.8 \\
\hline
\end{tabular}
\vspace{0.0cm} 
\caption{Ablation studies on Waymo validation set.}
\label{tab:3}
\end{table}

\noindent\textbf{Ablation Study.}
As shown in Table \ref{tab:3}, we compare the CenterPoint \cite{centerpoint}, Motion Transformer, Motion-Guided Feature Aggregation, and Dual Correlation Weighting Module sequentially for feature fusion structure, and we can see that CenterPoint is difficult to model multi-frame features. Meanwhile, modeling features solely through a Transformer can be challenging. The proposed methods MGFA and DCWM offer a significant enhancement in APH by 2-3\% through the incorporation of dynamic priors into the Transformer models.

\begin{table}
\centering
\fontsize{8}{9}\selectfont
\begin{tabular}{ccccccc}
\hline
\textbf{Experiment Number} & \textbf{Time} & \textbf{Veh. L2 APH} & \textbf{Ped. L2 APH} & \textbf{Cyl. L2 APH} \\
\hline
1 & $t$ & 62.13 & 60.91 & 61.16 \\
2 & $t-1,t$ & 63.41 & 58.17 & 62.37 \\
3 & $t-2,t-1,t$ & 63.46 & 58.62 & 63.89 \\
4 & $t-1,t,t+1$ & \bfseries 65.88 & 61.10 & \bfseries 68.80 \\
5 & $t-2,t-1,t,t+1,t+2$ & 65.73 & \bfseries 61.13 & 67.56 \\
\hline
\end{tabular}
\caption{Ablation study of the frame fusion effects on Waymo validation set.}
\label{tab:4}
\end{table}

 Since our task is a multi-frame fusion strategy, we need to consider the number of frames to be used. In Table \ref{tab:4}, we compare the effects of multi-frame fusion including single-frame, past-frame fusion, and past-future fusion. 
 In summary, we can draw three key conclusions. Firstly, the fusion of cross-frame, as seen (EXP. 1,2, and 4), significantly contributes to detection results.
 Secondly, using too many frames (EXP. 5) not only increases memory requirements but also hampers model convergence. The main reason this conclusion differs from MSF \cite{he2023msf} is that we use feature-level temporal fusion, whereas excessive attention stacking can hinder target characterization. 
 Lastly, relying solely on past frames limits the model's understanding of the scene's geometry (EXP. 3 and 4). Incorporating both past and future frames provides a more comprehensive context for improved performance.

\begin{table}[ht]
\begin{minipage}[t]{0.6\linewidth}
\centering
\footnotesize
\begin{tabular}{ccc}
\hline
\textbf{Motion Feature} & \textbf{Cyl. L2 APH} & \textbf{Veh. L2 APH} \\
\hline
pre2cur & 66.51 & 65.72 \\
fut2cur & 66.47 & 65.78 \\
cur2pre & 66.13 & 65.7 \\
cur2fut & 66.21 & 65.72 \\
cur2pre + cur2fut & 66.57 & 65.83 \\
pre2cur + fut2cur & \bfseries 68.80 & \bfseries 65.88 \\
\hline
\end{tabular}
\caption{Ablation study of motion-based heatmap \newline
feature selections on Waymo validation set.}
\label{tab:5}
\end{minipage}%
\begin{minipage}[t]{0.4\linewidth}
\centering
\footnotesize
\begin{tabular}{ccc}
\hline
\textbf{Fusion Method} & \textbf{NDS} & \textbf{mAP} \\
\hline
Concatenate & 66.79 & 58.13 \\
Attention \cite{vit} & 65.37 & 58.99 \\
Spatial Fusion \cite{centerformer} & 67.13 & 60.31 \\
DCWM & \bfseries 68.32 & \bfseries 63.77 \\
\hline
\end{tabular}
\caption{Ablation study of different feature fusion strategies  on Waymo validation set.}
\label{tab:6}
\end{minipage}
\end{table}

We select the motion feature as shown in Table \ref{tab:5}, it fuses the information of object motion and encodes its features according to trajectory predictions. 
However, we find the feature observed at different times does not have much effect on the metrics. 
It can be concluded that the trajectory feature predicted by the future and the past for the present works best and is the most logical. 
For multiple frames feature map fusion, we sequentially compare the following schemes, concatenate, attention, and spatial-aware attention which are mentioned in CenterFormer \cite{centerformer} and our proposed DCWM in Table \ref{tab:6}. We can discern that directly employing attention could hinder model learning, potentially yielding inferior results compared to concatenation and spatial fusion. However, our proposed Dual Correlation Weighting Module effectively fuses multiple frames and brings more pronounced enhancements.

\noindent\section{Counclusion}
\label{sec:intro}
Addressing the challenge of detecting sparse and occluded long-range LiDAR point clouds, we introduce LiSTM, a motion-based spatial-temporal fusion 3D point cloud detector. It leverages well-designed motion features and motion-guided feature fusion to enhance detection performance on Waymo and nuScenes datasets. 
In future work, we will focus on developing an end-to-end motion generator and exploring sparse feature representations.

\noindent\section*{Appendix}
\noindent\textbf{Computational Efficiency.}
We acknowledge that some reviewers have raised concerns regarding the computational resources. To address this, we compare CenterPoint, PVRCNN, MSF, and LiSTM. Despite LiSTM having significantly larger model parameters, its actual FPS is comparable to that of CenterPoint \cite{centerpoint}. Moreover, LiSTM demonstrates a nearly 50\% speed improvement over PV-RCNN++ \cite{pvrcnn++} while consuming less memory and operating more efficiently than MSF \cite{he2023msf}. This performance advantage primarily stems from our use of sparse feature operations and shared networks, which eliminate the need for computationally intensive processes such as multi-frame splicing and resampling.

\begin{table}[h]
\centering
\fontsize{7.5}{8.5}\selectfont
\begin{tabular}{@{}lcccc@{}}
\hline
\textbf{Model} & \textbf{Model Parameter} & \textbf{Memory cost} & \textbf{FPS} \\
\hline
CenterPoint \cite{centerpoint} & 7758811 & 2464 MiB & 5.68 it/s \\
PV-RCNN++ \cite{pvrcnn++} & 13073505 & 3918 MiB & 3.75 it/s \\
MSF \cite{he2023msf} & 15661651 & 6684 MiB & 4.58 it/s \\
LiSTM & 17592422 & 4400 MiB & 5.26 it/s \\
\hline
\end{tabular}
\caption{Computational efficiency}
\label{tab:7}
\end{table}

\noindent\textbf{Point-Trajectory Model Analysis and Performance Comparison.}
Taking MSF \cite{he2023msf} as an example, it enhances temporal features at the input level in two stages. In contrast, our approach targets implicit features, allowing for more efficient parallel computation and improved resource utilization. Unlike MSF's ROI sampling on point clouds, our method constructs a BEV heatmap, significantly boosting performance for larger targets like Vel (6m) and Cly (2m). However, for smaller targets like Ped (0.5m), even minor deviations can reduce performance, leading to lower results compared to MSF.

\noindent\textbf{Lack Related Work on The Motion Estimation Model.} 
Works \cite{stark,sttracker,DMT} have proposed learnable SOT models and we will try to complete the end-to-end model in this direction in the future. However, this class of methods requires significant computational resources and is not well-suited for multiple target detectors. Therefore, our proposed strategy is to use a simple linear Kalman model for target trajectory prediction, which characterizes target motion a priori without the need for learnable parameters or GPU resources.

\noindent\textbf{Total Waymo Evaluation.} 
The model validation results for Waymo's full training dataset are shown below, focusing on a comparison between CenterPoint \cite{centerpoint} and PVRCNN++ \cite{pvrcnn++}.
\begin{table}[H] 
\centering
\fontsize{6.6}{7.5}\selectfont
\begin{tabular}{@{\extracolsep{\fill}}l|cc|cc|cc}
\hline
\textbf{Model} & \multicolumn{2}{c|}{\textbf{Vehicle (AP/APH)$\uparrow$}} & \multicolumn{2}{c|}{\textbf{Pedestrian (AP/APH)$\uparrow$}} & \multicolumn{2}{c}{\textbf{Cyclist (AP/APH)$\uparrow$}} \\
 & \textbf{L1} & \textbf{L2} & \textbf{L1} & \textbf{L2} & \textbf{L1} & \textbf{L2} \\
\hline
CenterPoint \cite{centerpoint} & 72.64 / 72.10  & 64.57 / 64.07  & 74.53 / 68.36 & 66.50 / 60.84 & 71.14 / 69.91 & 68.56 / 67.37 \\
PV-RCNN++ \cite{pvrcnn++} & 77.80 / 77.34 & 69.43 / 69.01 & 80.00 / 73.94 & 71.62 / 65.97 & 72.43/ 71.35 & 69.79 / 68.74 \\
LiSTM & \bfseries 78.91 / \bfseries 78.31 & \bfseries 70.64 / \bfseries 70.10 & \bfseries 80.79 / \bfseries 75.01 & \bfseries 72.16 / \bfseries 66.87 & \bfseries 74.42 / \bfseries 73.33 & \bfseries 71.84 / \bfseries 70.79 \\
\hline
\end{tabular}
\caption{Quantative comparison on Waymo validation set.}
\label{tab:8}
\end{table}

\noindent\textbf{Long Distance Perception.} 
The LiSTM architecture leverages continuous frames and motion priors to enhance performance, particularly for long-range detection. In our evaluation with the Waymo dataset, which covers a 75m radius horizontally and vertically, we use three distance thresholds to metric. Results show that LiSTM outperforms the baseline by an average of 5 points in the 25m to 75m range. Even beyond this range, where the point cloud is mostly filtered out, LiSTM metrics remain somewhat elevated compared to the baseline.

\begin{table}[H]
\centering
\fontsize{6.0}{7.0}\selectfont
\begin{tabular}{l|ccc|ccc|ccc}
\hline
\textbf{Model} & \multicolumn{3}{c|}{\textbf{25m away mAP$\uparrow$}} & \multicolumn{3}{c|}{\textbf{50m-75m mAP$\uparrow$}} & \multicolumn{3}{c}{\textbf{75m away mAP$\uparrow$}} \\
& \textbf{Vehicle} & \textbf{Pedestrians} & \textbf{Cyclist} & \textbf{Vehicle} & \textbf{Pedestrians} & \textbf{Cyclist} & \textbf{Vehicle} & \textbf{Pedestrians} & \textbf{Cyclist} \\
\hline
CenterPoint \cite{centerpoint} & 58.80 & 63.12 & 61.37 & 41.82 & 54.00 & 50.50 & 11.46 & 16.30 & 14.82 \\
LiSTM & \bfseries 64.14 & \bfseries 68.05 & \bfseries 65.25 & \bfseries 46.31 & \bfseries 57.51 & \bfseries 53.87 & \bfseries 12.89 & \bfseries 17.25 & \bfseries 15.16 \\
\hline
\end{tabular}
\caption{Long distance perception metric on the Waymo validation set.}
\label{tab:9}
\end{table}

\bibliography{egbib}

\begin{thebibliography}{38}
\providecommand{\natexlab}[1]{#1}
\providecommand{\url}[1]{\texttt{#1}}
\expandafter\ifx\csname urlstyle\endcsname\relax
  \providecommand{\doi}[1]{doi: #1}\else
  \providecommand{\doi}{doi: \begingroup \urlstyle{rm}\Url}\fi

\bibitem[Caesar et~al.(2020)Caesar, Bankiti, Lang, Vora, Liong, Xu, Krishnan, Pan, Baldan, and Beijbom]{nuscenes}
Holger Caesar, Varun Bankiti, Alex~H Lang, Sourabh Vora, Venice~Erin Liong, Qiang Xu, Anush Krishnan, Yu~Pan, Giancarlo Baldan, and Oscar Beijbom.
\newblock nuscenes: A multimodal dataset for autonomous driving.
\newblock In \emph{Proceedings of the IEEE/CVF Conference on Computer Vision and Pattern Recognition}, pages 11621--11631, 2020.

\bibitem[Calvo et~al.(2023)Calvo, Taveira, Kahl, Gustafsson, Larsson, and Tonderski]{timepillars}
Ernesto~Lozano Calvo, Bernardo Taveira, Fredrik Kahl, Niklas Gustafsson, Jonathan Larsson, and Adam Tonderski.
\newblock Timepillars: Temporally-recurrent 3d lidar object detection.
\newblock \emph{arXiv preprint arXiv:2312.17260}, 2023.

\bibitem[Chen et~al.(2022)Chen, Shi, Zhu, Cheung, Xu, and Li]{mppnet}
Xuesong Chen, Shaoshuai Shi, Benjin Zhu, Ka~Chun Cheung, Hang Xu, and Hongsheng Li.
\newblock Mppnet: Multi-frame feature intertwining with proxy points for 3d temporal object detection.
\newblock In \emph{European Conference on Computer Vision}, pages 680--697. Springer, 2022.

\bibitem[Chen et~al.(2023)Chen, Liu, Zhang, Qi, and Jia]{voxelnext}
Yukang Chen, Jianhui Liu, Xiangyu Zhang, Xiaojuan Qi, and Jiaya Jia.
\newblock Voxelnext: Fully sparse voxelnet for 3d object detection and tracking.
\newblock In \emph{Proceedings of the IEEE/CVF Conference on Computer Vision and Pattern Recognition}, pages 21674--21683, 2023.

\bibitem[Cui et~al.(2023)Cui, Li, and Fang]{sttracker}
Yubo Cui, Zhiheng Li, and Zheng Fang.
\newblock Sttracker: Spatio-temporal tracker for 3d single object tracking.
\newblock \emph{IEEE Robotics and Automation Letters}, 2023.

\bibitem[Deng et~al.(2021)Deng, Shi, Li, Zhou, Zhang, and Li]{voxelrcnn}
Jiajun Deng, Shaoshuai Shi, Peiwei Li, Wengang Zhou, Yanyong Zhang, and Houqiang Li.
\newblock Voxel r-cnn: Towards high performance voxel-based 3d object detection.
\newblock In \emph{Proceedings of the AAAI Conference on Artificial Intelligence}, volume~35, pages 1201--1209, 2021.

\bibitem[Dosovitskiy et~al.(2020)Dosovitskiy, Beyer, Kolesnikov, Weissenborn, Zhai, Unterthiner, Dehghani, Minderer, Heigold, Gelly, et~al.]{vit}
Alexey Dosovitskiy, Lucas Beyer, Alexander Kolesnikov, Dirk Weissenborn, Xiaohua Zhai, Thomas Unterthiner, Mostafa Dehghani, Matthias Minderer, Georg Heigold, Sylvain Gelly, et~al.
\newblock An image is worth 16x16 words: Transformers for image recognition at scale.
\newblock \emph{arXiv preprint arXiv:2010.11929}, 2020.

\bibitem[He et~al.(2023)He, Li, Zhang, Li, and Zhang]{he2023msf}
Chenhang He, Ruihuang Li, Yabin Zhang, Shuai Li, and Lei Zhang.
\newblock Msf: Motion-guided sequential fusion for efficient 3d object detection from point cloud sequences.
\newblock In \emph{Proceedings of the IEEE/CVF Conference on Computer Vision and Pattern Recognition}, pages 5196--5205, 2023.

\bibitem[Huang et~al.(2024)Huang, Lyu, Yang, and Tsai]{ptt}
Kuan-Chih Huang, Weijie Lyu, Ming-Hsuan Yang, and Yi-Hsuan Tsai.
\newblock Ptt: Point-trajectory transformer for efficient temporal 3d object detection.
\newblock In \emph{Proceedings of the IEEE/CVF Conference on Computer Vision and Pattern Recognition}, pages 14938--14947, 2024.

\bibitem[Kim et~al.(2021)Kim, O{\v{s}}ep, and Leal-Taix{\'e}]{eagermot}
Aleksandr Kim, Aljo{\v{s}}a O{\v{s}}ep, and Laura Leal-Taix{\'e}.
\newblock Eagermot: 3d multi-object tracking via sensor fusion.
\newblock In \emph{2021 IEEE International Conference on Robotics and Automation}, pages 11315--11321. IEEE, 2021.

\bibitem[Lang et~al.(2019)Lang, Vora, Caesar, Zhou, Yang, and Beijbom]{pointpillars}
Alex~H Lang, Sourabh Vora, Holger Caesar, Lubing Zhou, Jiong Yang, and Oscar Beijbom.
\newblock Pointpillars: Fast encoders for object detection from point clouds.
\newblock In \emph{Proceedings of the IEEE/CVF Conference on Computer Vision and Pattern Recognition}, pages 12697--12705, 2019.

\bibitem[Law and Deng(2018)]{cornnet}
Hei Law and Jia Deng.
\newblock Cornernet: Detecting objects as paired keypoints.
\newblock In \emph{European Conference on Computer Vision}, pages 734--750, 2018.

\bibitem[Li et~al.(2023{\natexlab{a}})Li, Qi, Zhou, Liu, and Anguelov]{modar}
Yingwei Li, Charles~R Qi, Yin Zhou, Chenxi Liu, and Dragomir Anguelov.
\newblock Modar: Using motion forecasting for 3d object detection in point cloud sequences.
\newblock In \emph{Proceedings of the IEEE/CVF Conference on Computer Vision and Pattern Recognition}, pages 9329--9339, 2023{\natexlab{a}}.

\bibitem[Li et~al.(2023{\natexlab{b}})Li, Ge, Yu, Yang, Wang, Shi, Sun, and Li]{bevdepth}
Yinhao Li, Zheng Ge, Guanyi Yu, Jinrong Yang, Zengran Wang, Yukang Shi, Jianjian Sun, and Zeming Li.
\newblock Bevdepth: Acquisition of reliable depth for multi-view 3d object detection.
\newblock In \emph{Proceedings of the AAAI Conference on Artificial Intelligence}, volume~37, pages 1477--1485, 2023{\natexlab{b}}.

\bibitem[Li et~al.(2022)Li, Wang, Li, Xie, Sima, Lu, Qiao, and Dai]{bevformer}
Zhiqi Li, Wenhai Wang, Hongyang Li, Enze Xie, Chonghao Sima, Tong Lu, Yu~Qiao, and Jifeng Dai.
\newblock Bevformer: Learning bird’s-eye-view representation from multi-camera images via spatiotemporal transformers.
\newblock In \emph{European Conference on Computer Vision}, pages 1--18. Springer, 2022.

\bibitem[Liu et~al.(2016)Liu, Anguelov, Erhan, Szegedy, Reed, Fu, and Berg]{ssd}
Wei Liu, Dragomir Anguelov, Dumitru Erhan, Christian Szegedy, Scott Reed, Cheng-Yang Fu, and Alexander~C Berg.
\newblock Ssd: Single shot multibox detector.
\newblock In \emph{European Conference on Computer Vision}, pages 21--37. Springer, 2016.

\bibitem[Philion and Fidler(2020)]{lss}
Jonah Philion and Sanja Fidler.
\newblock Lift, splat, shoot: Encoding images from arbitrary camera rigs by implicitly unprojecting to 3d.
\newblock In \emph{European Conference on Computer Vision}, pages 194--210. Springer, 2020.

\bibitem[Qi et~al.(2017)Qi, Su, Mo, and Guibas]{pointnet}
Charles~R Qi, Hao Su, Kaichun Mo, and Leonidas~J Guibas.
\newblock Pointnet: Deep learning on point sets for 3d classification and segmentation.
\newblock In \emph{Proceedings of the IEEE/CVF Conference on Computer Vision and Pattern Recognition}, pages 652--660, 2017.

\bibitem[Shi et~al.(2022)Shi, Li, and Ma]{pillarnet}
Guangsheng Shi, Ruifeng Li, and Chao Ma.
\newblock Pillarnet: Real-time and high-performance pillar-based 3d object detection.
\newblock In \emph{European Conference on Computer Vision}, pages 35--52. Springer, 2022.

\bibitem[Shi et~al.(2019{\natexlab{a}})Shi, Wang, and Li]{pointrcnn}
Shaoshuai Shi, Xiaogang Wang, and Hongsheng Li.
\newblock Pointrcnn: 3d object proposal generation and detection from point cloud.
\newblock In \emph{Proceedings of the IEEE/CVF Conference on Computer Vision and Pattern Recognition}, pages 770--779, 2019{\natexlab{a}}.

\bibitem[Shi et~al.(2019{\natexlab{b}})Shi, Wang, Wang, and Li]{PARTA2}
Shaoshuai Shi, Zhe Wang, Xiaogang Wang, and Hongsheng Li.
\newblock Part-a2 net: 3d part-aware and aggregation neural network for object detection from point cloud.
\newblock \emph{arXiv preprint arXiv:1907.03670}, 2\penalty0 (3), 2019{\natexlab{b}}.

\bibitem[Shi et~al.(2020)Shi, Guo, Jiang, Wang, Shi, Wang, and Li]{pvrcnn}
Shaoshuai Shi, Chaoxu Guo, Li~Jiang, Zhe Wang, Jianping Shi, Xiaogang Wang, and Hongsheng Li.
\newblock Pv-rcnn: Point-voxel feature set abstraction for 3d object detection.
\newblock In \emph{Proceedings of the IEEE/CVF Conference on Computer Vision and Pattern Recognition}, pages 10529--10538, 2020.

\bibitem[Shi et~al.(2023)Shi, Jiang, Deng, Wang, Guo, Shi, Wang, and Li]{pvrcnn++}
Shaoshuai Shi, Li~Jiang, Jiajun Deng, Zhe Wang, Chaoxu Guo, Jianping Shi, Xiaogang Wang, and Hongsheng Li.
\newblock Pv-rcnn++: Point-voxel feature set abstraction with local vector representation for 3d object detection.
\newblock \emph{International Journal of Computer Vision}, 131\penalty0 (2):\penalty0 531--551, 2023.

\bibitem[Sun et~al.(2020)Sun, Kretzschmar, Dotiwalla, Chouard, Patnaik, Tsui, Guo, Zhou, Chai, Caine, et~al.]{waymodataset}
Pei Sun, Henrik Kretzschmar, Xerxes Dotiwalla, Aurelien Chouard, Vijaysai Patnaik, Paul Tsui, James Guo, Yin Zhou, Yuning Chai, Benjamin Caine, et~al.
\newblock Scalability in perception for autonomous driving: Waymo open dataset, 2020.

\bibitem[Team(2020)]{openpcdet}
OpenPCDet~Development Team.
\newblock Openpcdet: An open-source toolbox for 3d object detection from point clouds.
\newblock \url{GitHub - open-mmlab/OpenPCDet: OpenPCDet Toolbox for LiDAR-based 3D Object Detection.}, 2020.

\bibitem[Tian et~al.(2019)Tian, Yang, Wang, Wang, Li, and Liang]{yolov3}
Yunong Tian, Guodong Yang, Zhe Wang, Hao Wang, En~Li, and Zize Liang.
\newblock Apple detection during different growth stages in orchards using the improved yolo-v3 model.
\newblock \emph{Computers and Electronics in Agriculture}, 157:\penalty0 417--426, 2019.

\bibitem[Vaswani et~al.(2017)Vaswani, Shazeer, Parmar, Uszkoreit, Jones, Gomez, Kaiser, and Polosukhin]{attention}
Ashish Vaswani, Noam Shazeer, Niki Parmar, Jakob Uszkoreit, Llion Jones, Aidan~N Gomez, {\L}ukasz Kaiser, and Illia Polosukhin.
\newblock Attention is all you need.
\newblock \emph{Advances in Neural Information Processing Systems}, 30, 2017.

\bibitem[Wang et~al.(2023)Wang, Liu, Wang, Li, and Zhang]{streampetr}
Shihao Wang, Yingfei Liu, Tiancai Wang, Ying Li, and Xiangyu Zhang.
\newblock Exploring object-centric temporal modeling for efficient multi-view 3d object detection.
\newblock In \emph{Proceedings of the IEEE/CVF International Conference on Computer Vision}, pages 3621--3631, 2023.

\bibitem[Xia et~al.(2023)Xia, Wu, Li, Chan, and Stilla]{DMT}
Yan Xia, Qiangqiang Wu, Wei Li, Antoni~B Chan, and Uwe Stilla.
\newblock A lightweight and detector-free 3d single object tracker on point clouds.
\newblock \emph{IEEE Transactions on Intelligent Transportation Systems}, 24\penalty0 (5):\penalty0 5543--5554, 2023.

\bibitem[Yan et~al.(2021)Yan, Peng, Fu, Wang, and Lu]{stark}
Bin Yan, Houwen Peng, Jianlong Fu, Dong Wang, and Huchuan Lu.
\newblock Learning spatio-temporal transformer for visual tracking.
\newblock In \emph{Proceedings of the IEEE/CVF International Conference on Computer Vision}, pages 10448--10457, 2021.

\bibitem[Yan et~al.(2018)Yan, Mao, and Li]{second}
Yan Yan, Yuxing Mao, and Bo~Li.
\newblock Second: Sparsely embedded convolutional detection.
\newblock \emph{Sensors}, 18\penalty0 (10):\penalty0 3337, 2018.

\bibitem[Yang et~al.(2021)Yang, Zhou, Chen, and Ngiam]{3dman}
Zetong Yang, Yin Zhou, Zhifeng Chen, and Jiquan Ngiam.
\newblock 3d-man: 3d multi-frame attention network for object detection.
\newblock In \emph{Proceedings of the IEEE/CVF Conference on Computer Vision and Pattern Recognition}, pages 1863--1872, 2021.

\bibitem[Yin et~al.(2020)Yin, Shen, Guan, Zhou, and Yang]{convgru}
Junbo Yin, Jianbing Shen, Chenye Guan, Dingfu Zhou, and Ruigang Yang.
\newblock Lidar-based online 3d video object detection with graph-based message passing and spatiotemporal transformer attention.
\newblock In \emph{Proceedings of the IEEE/CVF Conference on Computer Vision and Pattern Recognition}, pages 11495--11504, 2020.

\bibitem[Yin et~al.(2021)Yin, Zhou, and Krahenbuhl]{centerpoint}
Tianwei Yin, Xingyi Zhou, and Philipp Krahenbuhl.
\newblock Center-based 3d object detection and tracking.
\newblock In \emph{Proceedings of the IEEE/CVF Conference on Computer Vision and Pattern Recognition}, pages 11784--11793, 2021.

\bibitem[Zhao et~al.(2024)Zhao, Heng, Wang, Gao, Liu, Yao, Chen, and Cai]{3dlane}
Runkai Zhao, Yuwen Heng, Heng Wang, Yuanda Gao, Shilei Liu, Changhao Yao, Jiawen Chen, and Weidong Cai.
\newblock Advancements in 3d lane detection using lidar point clouds: From data collection to model development.
\newblock In \emph{2024 IEEE International Conference on Robotics and Automation}, pages 5382--5388. IEEE, 2024.

\bibitem[Zhou et~al.(2019)Zhou, Wang, and Kr{\"a}henb{\"u}hl]{objectspoint}
Xingyi Zhou, Dequan Wang, and Philipp Kr{\"a}henb{\"u}hl.
\newblock Objects as points.
\newblock \emph{arXiv preprint arXiv:1904.07850}, 2019.

\bibitem[Zhou and Tuzel(2018)]{voxelnet}
Yin Zhou and Oncel Tuzel.
\newblock Voxelnet: End-to-end learning for point cloud based 3d object detection.
\newblock In \emph{Proceedings of the IEEE/CVF Conference on Computer Vision and Pattern Recognition}, pages 4490--4499, 2018.

\bibitem[Zhou et~al.(2022)Zhou, Zhao, Wang, Wang, and Foroosh]{centerformer}
Zixiang Zhou, Xiangchen Zhao, Yu~Wang, Panqu Wang, and Hassan Foroosh.
\newblock Centerformer: Center-based transformer for 3d object detection.
\newblock In \emph{European Conference on Computer Vision}, pages 496--513. Springer, 2022.

\end{thebibliography}

\end{document}